\documentclass{article}

\usepackage{graphicx,color,amssymb,amsmath}
\usepackage{booktabs}
\usepackage{url}
\usepackage[authoryear]{natbib}
\usepackage{hyperref}
\hypersetup{
    colorlinks=true,
    linkcolor=blue,
    filecolor=blue,      
    urlcolor=blue,
    citecolor=blue,
}
\usepackage{authblk}

\title{An Accurate and Fully-Automated Ensemble Model for Weekly Time Series Forecasting}

\author[1,*]{Rakshitha Godahewa}
\author[1]{Christoph Bergmeir}
\author[1]{Geoffrey I. Webb}
\author[2]{Pablo Montero-Manso}

\affil[1]{Department of Data Science and Artificial Intelligence\\
    Monash University, Australia.}
\affil[2]{University of Sydney, Australia}
\affil[*]{Corresponding Author: rakshitha.godahewa@monash.edu}
	
\date{}

\begin{document}

\maketitle

\begin{abstract}
Many businesses and industries require accurate forecasts for weekly time series nowadays. However, the forecasting literature does not currently provide easy-to-use, automatic, reproducible and accurate approaches dedicated to this task. We propose a forecasting method in this domain to fill this gap, leveraging state-of-the-art forecasting techniques, such as forecast combination, meta-learning, and global modelling. We consider different meta-learning architectures, algorithms, and base model pools. Based on all considered model variants, we propose to use a stacking approach with lasso regression which optimally combines the forecasts of four base models: a global Recurrent Neural Network (RNN) model, Theta, Trigonometric Box–Cox ARMA Trend Seasonal (TBATS), and Dynamic Harmonic Regression ARIMA (DHR-ARIMA), as it shows the overall best performance across seven experimental weekly datasets on four evaluation metrics. Our proposed method also consistently outperforms a set of benchmarks and state-of-the-art weekly forecasting models by a considerable margin with statistical significance. Our method can produce the most accurate forecasts, in terms of mean sMAPE, for the M4 weekly dataset among all benchmarks and all original competition participants.
\end{abstract}

%\begin{keyword}
%Weekly Forecasting \sep Global Models \sep Ensembling \sep Meta-Learning
%\end{keyword}

\section{Introduction}
\label{sec:intro}

Even though weekly time series are an important type of series in many business contexts, forecasters have traditionally focused more on series with lower or higher granularity. The M3 and M4 competitions \citep{makridakis_2000_m3, makridakis_2018_m4} were dominated by monthly series, besides strong presences of quarterly and yearly series. On the other hand, daily and sub-daily series also gain considerable attention as they often show multiple seasonalities which makes them an interesting field of study \citep{bandara_2019_msnet, Hara-Wild_2018_fasster}. 

Weekly time series generally have only a single seasonality with a long cycle, i.e., a yearly seasonality with a seasonal cycle of approximate length 52.18, which is a large and a non-integer value compared with the other commonly considered seasonalities such as quarterly and monthly. On the other hand, when an integer seasonality is required, there are two possible values, 52 and 53, depending on the year. Selection between these integer and non-integer values for the seasonal period depends on the situation. This particular seasonality needs to be properly accounted for in the modelling. Furthermore, due to the long seasonal cycle, weekly series often do not have data spanning over two full periods (two years) where this has consequences for the modelling process.

Even though modelling weekly time series is not as popular a use case as modelling quarterly and monthly series, weekly series are very important in practice. As storing and processing large amounts of data is almost trivial nowadays, and business processes become more and more automated, many businesses that have traditionally operated on quarterly or monthly bases are now operating on a weekly time scale, and consequently, practitioners are often interested in forecasting product sales on a weekly basis. 

Consequently, in the M4, the best-performing methods in the weekly category were quite different to the overall winning methods. The method that won the M4-weekly category was a solution based on the commercial Forecast Pro software, discussed in detail by \cite{goodrich_2020_forecast_pro}, where notably those authors also argue that the M4 in general is not very representative for a usual business situation but ``the weekly data are perhaps the most similar to business data.''
Though the method of \citet{goodrich_2020_forecast_pro} is very accurate, it requires a significant amount of manual intervention and thorough domain knowledge to retrieve good forecasts, and it is based on a proprietary software. 

On the other hand, the methods that performed overall well in the competition popularised important methodological innovations in the forecasting field. Namely 1) forecast combination (ensembling) with meta-learning, as used in the second-winning method from \citet{pablo_2020_fforma} that combines the forecasts of a set of base algorithms, and 2) the use of global forecasting models \citep{januschowski_2020_criteria}, as in the winning method from \citet{smyl_2020_hybrid}, which is a global forecasting method (that also uses ensembling at different stages of the algorithm).

The main contribution of this paper is to introduce an accurate, simple, fully-automated, publicly available, and reproducible weekly forecasting method using state-of-the-art forecasting practices of global forecasting models and forecast combination. In particular, we propose an automated meta-learning ensemble forecasting model for weekly time series forecasting. We analyse different base model pools where the forecasts of the base models are combined using different meta-learning architectures that are based on features, such as the architecture of \citet{pablo_2020_fforma}, and on stacking \citep{wolpert_1992_stacked}. Furthermore, we use three meta-learning algorithms: linear regression, lasso regression \citep{tibshirani_1994_regression}, and eXtreme Gradient Boosting \citep[XGBoost,][]{chen_xgboost_2016}. Based on all considered model variants, the stacking approach which combines the forecasts of four base models: a globally trained Recurrent Neural Network \citep[RNN,][]{hewamalage_2019_recurrent} and three univariate forecasting models: Theta \citep{assimakopoulos_2000_theta}, Trigonometric Box-Cox ARMA Trend Seasonal \citep[TBATS, ][]{livera_2011_tbats}, and Dynamic Harmonic Regression Auto-Regressive Integrated Moving Average \citep[DHR-ARIMA,][]{hyndman_2018_fpp}, using lasso regression shows the overall best performance.
 The sub-models that we use in this best performing proposed method include both global and local forecasting models that are especially suitable for weekly forecasting where the most suitable sub-models are automatically selected, and their forecasts are combined using the lasso regression meta-learning algorithm. Hence, this weekly forecasting method in particular outperforms a set of state-of-the-art weekly forecasting benchmarks with statistical significance across seven experimental datasets on four error metrics. Furthermore, our method produces the most accurate forecasts for the M4 weekly dataset based on the mean of the symmetric Mean Absolute Percentage Error (sMAPE), compared with all original competition participants. Our method is fully-automated and domain knowledge is not required to obtain the forecasts. Since our method does not use any seasonal features extracted from the series, it is applicable to any weekly dataset irrespective of the series length. 
In this study, we use publicly available benchmark datasets that are representative for many real-world applications. However, as their size is limited, when scaling the method to larger datasets, some of our recommendations and modelling choices within the framework, for example the use of lasso regression, may need to be revisited.
All implementations related to the method are publicly available at: \url{https://github.com/rakshitha123/WeeklyForecasting}.
Out of our seven benchmark datasets, five datasets are created by aggregating series with higher granularities. We make all the aggregated weekly datasets publicly available for further research use\footnote{\url{https://github.com/rakshitha123/WeeklyForecasting/tree/master/datasets}}.   

The remainder of this paper is organised as follows: Section \ref{sec:relw}  reviews  the  related  work. Section \ref{sec:methodology} explains the main components of our analysis including the details of the base forecasting model pools, meta-learning architectures and algorithms used for aggregating forecasts. Section \ref{sec:experiments} explains our experimental framework. Section \ref{sec:eval_accuracy} presents the model evaluation results. Finally, Section \ref{sec:conclusion} concludes the paper and discusses possible future research.

\section{Related Work}
\label{sec:relw}

In the following, we discuss the related literature in the field of weekly time series forecasting and summarise the state of the art in the fields of meta-learning for forecast combination, and global modelling.

\subsection{Weekly Forecasting}
\label{sec:weekly_forecasting_literature}

The main challenge when forecasting weekly data is the long and non-integer yearly seasonality.
The most popular general forecasting techniques Exponential Smoothing State Space Models \citep[ETS, ][]{hyndman_2008_ets} and Auto-Regressive Integrated Moving Average \citep[ARIMA, ][]{box_1990_arima} are not able to deal well with such a long seasonal cycle, and hence, their applicability in weekly forecasting is limited \citep{hyndman_2018_fpp}.
There are four main ways in the literature to deal with the seasonality in weekly data. The seasonality 1) can be neglected and a non-seasonal model be built, it 2) can be addressed with seasonal lags, it 3) can be addressed with seasonal indicator variables such as Fourier terms or seasonal dummy variables, or 4) the series can be deseasonalised before forecasting.

Building non-seasonal models can be a good option if the yearly seasonality is not strong or if less than a full year of data is available. In that case, non-seasonal standard models such as ETS, ARIMA, or Theta \citep{assimakopoulos_2000_theta} can be fitted. For example, \cite{Padt2017Optical} use simple exponential smoothing in such a situation. Also commonly employed in the literature are other (non-seasonal) non-linear autoregressive models, e.g., from a machine learning domain.
\citet{qaness_2020_influenza} propose an improved version of an Adaptive Neuro-Fuzzy Inference System (ANFIS) to forecast the weekly confirmed influenza cases in China and the USA which can be used to support health policy-making. Researchers have also proposed models for weekly load forecasting \citep{barakat_1998_load}, weekly crude oil forecasting \citep{oussalah_2018_oil} and weekly groundwater level forecasting \citep{mohanty_2015_groundwater} where Neural Networks (NN) and regression models are heavily used with the proposed approaches. 

Seasonality can also be modelled by incorporating seasonal lags. For example, \citet{landeras_2009_evapotranspiration} use seasonal ARIMA and NN models to obtain weekly evapotranspiration forecasts which are then used in planning and designing water resource systems. 
However, this approach is hindered by the long seasonal cycle and the non-integer length of the cycle in weekly series, and a lag selection mechanism may be necessary to obtain good results.

Fourier terms are also commonly employed nowadays with machine learning models more broadly to model other types of seasonalities \citep{bandara_2019_msnet}. They are commonly used to model the long seasonal cycle of weekly data. TBATS \citep{livera_2011_tbats} and DHR-ARIMA are advocated as suitable methods to be used instead of ETS and ARIMA \citep{hyndman_2018_fpp} where both methods use Fourier terms to model the seasonality. 
\citet{pan_2017_hotel} use Autoregressive Moving Average eXogenous (ARMAX) models along with search engine queries, website traffic data and weather information to forecast weekly hotel occupancy for a particular destination. Those authors use Fourier terms and weekly dummy variables to model the seasonality.

Deseasonalisation is another possible approach to handle seasonality. \citet{guttormsen_1999_salmon} uses six models to forecast weekly prices for salmon where the forecasts can then be used to reduce the price fluctuation risks. Here, the time series are deseasonalised before forecasting. That author concludes that Classical Additive Decomposition (CAD) and Vector Auto Regression (VAR) models provide the best price forecasts. 

The winning approach of the M4 weekly competition proposed by \citet{goodrich_2020_forecast_pro} uses a set of experts which is chosen from a pool of baseline models (model families), mostly variations of exponential smoothing and ARIMA, to forecast each series. Furthermore, the baseline models include na\"ive2, seasonal na\"ive and dynamic regression models incorporating seasonal lags and Fourier terms to capture the seasonal effects. Short series that seem to be seasonal are customised to properly capture the seasonal effects. For each series, one or more models are selected from the pool of baseline models to obtain forecasts based on the series' characteristics. If more models are selected, then the approach uses an out-of-sample testing procedure to select the most appropriate base model to obtain forecasts for a given series. Further investigations are performed to identify the series where the selected experts are not adequate and finally, the forecasts of such series are modified using domain knowledge as required. Although this model is highly accurate, it requires considerable manual intervention during the forecasting process. 

\subsection{Meta-Learning for Forecast Combination}

Combining forecasts of several heterogeneous models is well-known to improve the forecasting accuracy on average over the individual models \citep{granger_1969_combination,timmermann_2006_forecast}, especially by reducing bias and/or variance \citep{wolpert_1992_stacked, cerqueira_2017_dynamic}. In the machine learning space, forecast combination is known as ensembling. Simple averaging is considered as one of the most efficient and accurate forecast combination methods which tends to be very competitive \citep{timmermann_2006_forecast}. Several weighted averaging combination methods have also been proposed \citep{sanchez_2008_adaptive, cerqueira_2017_dynamic}.

Recently, more sophisticated models for forecast combination often use meta-learning approaches. The second winning approach of the M4 forecasting competition, Feature-based Forecast Model Averaging  \citep[FFORMA,][]{pablo_2020_fforma}, as such an approach combines the forecasts provided by a set of base models using an optimal set of weights obtained using a meta-learner trained using series features (for details see Section \ref{sec:fforma_metalearner}). The third winning approach of the M4 forecasting competition proposed by \citet{PAWLIKOWSKI202093} also incorporates forecast combinations with meta-learning where the forecasts that are provided by a set of selected statistical models including ETS, ARIMA, Theta and Na\"ive are optimally combined for each series to obtain the final forecasts. Stacking \citep{wolpert_1992_stacked} is another widely used meta-learning approach in the forecast combination space \citep{Divina2018StackingEL, Khairalla2018ShortTermFF, ZHAI2018644}. It first trains a set of base models using a training set and then uses the predictions of the base models on the training set as inputs to train a meta-learner, which outputs the final predictions. Apart from the base model predictions, the meta-learner can have additional inputs such as features extracted from the series.

\subsection{Global Models in Forecasting}

Another relatively recent trend in forecasting, employed in the M4 successfully by the winning solution of \citet{smyl_2020_hybrid}, is the use of \emph{global forecasting models} \citep{januschowski_2020_criteria}.
Global models build a single model with a set of global parameters across many series. In contrast to local models, they are capable of learning the cross-series information during model training with a fewer amount of parameters. This type of models has been pioneered, e.g., by the works of \citet{duncan_2001_forecasting, trapero_2015_sales, smyl_2016_data, flunkert_2017_deepar, bandara_2020_clustering, pablo_2020_principles}. 
Many researchers have used RNNs in the context of global modelling \citep{smyl_2020_hybrid,hewamalage_2019_recurrent, bandara_2019_msnet, bandara_2019_sales, bandara_2020_clustering}. 

\bigskip

In this paper, we incorporate the capabilities of different meta-learning architectures that use both global and traditional univariate forecasting models to improve the accuracy of weekly time series forecasting without manual intervention.

\section{Methodology}
\label{sec:methodology}

\begin{figure}
\centering
 \includegraphics[width=\textwidth]{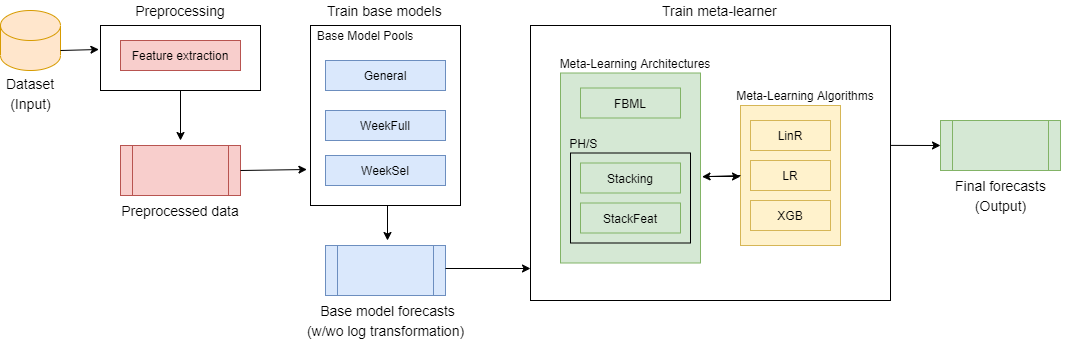}
 \caption{Overview of our modelling strategy together with the data preprocessing, base model pools, meta-learning architectures and meta-learning algorithms we consider for our study. Here, our goal is to find the best base model pool, meta-learning architecture and meta-learning algorithm to be used as the building blocks of an accurate ensemble weekly forecasting model.}~\label{fig:variants}
\end{figure}

The main purpose of our paper is to incorporate global modelling, forecast combinations and meta-learning to implement a fully-automated and accurate weekly forecasting model. This section details the base models and the meta-learning architectures, as well as the meta-learning algorithms that we use for aggregating the sub-model forecasts in our analysis.
Figure \ref{fig:variants} gives an overview of our modelling strategy together with the data preprocessing, base model pools, meta-learning architectures and meta-learning algorithms we consider for our study.

\subsection{Base Models}
\label{sec:base_models}

Forecast combination models contain a series of base models where the corresponding base-model forecasts are aggregated to obtain the final forecasts. 
We consider different base models for our analysis as detailed in the next sections.

\subsubsection{A General Base Model Pool}
\label{sec:fforma}

Forecast combinations are expected to produce better results when the base models are heterogeneous and produce relatively uncorrelated forecasts \citep{DEMENEZES2000190}.
Following these guidelines, an excellent starting point for a well-crafted and diverse set of base learners is the FFORMA method \citep{pablo_2020_fforma}. FFORMA is the second winning approach of the M4 forecasting competition. It optimally combines forecasts produced by nine sub-forecasting models: ETS, automated ARIMA (Auto.ARIMA), TBATS, Theta, na\"ive, sna\"ive, random walk with drift, a locally trained neural network (NNET-AR) and Seasonal and Trend decomposition using Loess with AR modeling of the seasonally adjusted series (STLM-AR). 
For our analysis, we consider all base models in FFORMA in their implementations in the R package \textit{forecast} \citep{hyndman_2008_automatic}. 

We note that FFORMA uses with ETS and Auto.ARIMA two base models that themselves have model selection capabilities and select a best model out of various sub-models, using mechanisms such as minimising a bias-corrected version of the Akaike Information Criterion \citep[AICc,][]{Akaike2011}. 
Another way of including these base models into a meta-learning framework (also see the approaches of \citet{goodrich_2020_forecast_pro} and \citet{KOLASSA2011238}), would be to use all the different model variants as base learners and use the meta-learner to select between the variants. However, while this is a possibility for ETS with a well-defined and relatively low amount of distinct models, for ARIMA models of different orders (or even ARIMA and ETS models with different coefficients) this leads to potentially a very high amount of base models. Therefore, we follow in our work the approach of FFORMA and opt for using any model selection procedures present in the base learners, under the assumption that they are crafted with expert knowledge of the base learners, and that this leads to a model pool of few strong and heterogeneous base models.

\subsubsection{Base Models for Weekly Forecasting}
\label{sec:suitability_weekly}

Out of the general base model pool identified above, some models are more suitable for weekly forecasting than others. 
The simple non-seasonal models Theta, na\"ive, and random walk with drift have limited suitability, as they do not take the seasonality into account. If they have a trend component, they will instead model the seasonality as a trend, which can lead to very bad accuracies. However, due to the long period of the seasonality, in practice often it may not be possible to model the seasonality, and if the seasonal pattern is not strong or slow-moving, ignoring it or modelling it through a trend may be acceptable. In the literature (Section \ref{sec:weekly_forecasting_literature}) there are many examples of non-seasonal models applied to weekly series with good results.
Next, sna\"ive and STLM-AR are simple methods that are able to model the seasonality in weekly data, though STLM-AR needs two full periods. 
TBATS \citep{livera_2011_tbats} is capable of dealing with long and non-integer periodic effects of time series which makes it one of the state-of-the-art methods used in weekly time series forecasting \citep{hyndman_2018_fpp}. TBATS models the seasonality using a Fourier-series-based trigonometric representation.
ETS \citep{hyndman_2008_ets}, as implemented in the \verb|ets()| function in the \textit{forecast} package in R \citep{hyndman_2008_automatic}, in the weekly forecasting context is a non-seasonal model as it does not handle seasonal periods greater than 24. 
A different implementation of ETS is available in the R package \textit{smooth} \citep{svetunkov_2020_smooth}. In contrast to ETS, this implementation handles seasonal cycles greater than 24 and hence, it can be used as a seasonal model in the weekly context. However, \cite{hyndman_2008_ets} argues that exponential smoothing has limitations if used with long seasonal cycles, and the limitation of \verb|ets()| to handle seasonal cycles only up to the length 24 is a design choice and not a limitation of the implementation. In line with this, we performed preliminary experiments with seasonal ETS as a base model, which did not provide promising results and therewith confirmed this assumption. Hence, we do not consider seasonal ETS and use ETS only in a non-seasonal version.
Auto.ARIMA \citep{hyndman_2008_automatic} is a state-of-the-art statistical benchmark, however, again, due to the long seasonal cycle of weekly series, it is considered not adequate for weekly series \citep{hyndman_2018_fpp}, and with DHR-ARIMA an alternative better suited for weekly time series exists, which is discussed in the following. 
Similarly, NNET-AR is a locally trained feed-forward NN and much better alternatives exist, namely we will use a globally trained RNNs, also discussed in the following.

\paragraph{Dynamic Harmonic Regression ARIMA} This technique is especially suitable to forecast series with long seasonal periods such as weekly series \citep{hyndman_2018_fpp}. The seasonality is considered to be fixed during modelling. Here, Fourier terms and ARMA errors are used to model the seasonality and short-term time series dynamics, respectively. 

Fourier terms are a set of sine and cosine pairs which are useful in modelling periodic effects in time series \citep{harvey_1993_structural}, especially to model the long seasonal periods presented in weekly data. The Fourier terms related to a particular time point of a series can be obtained using the formula shown in Equation \ref{eqn:fourier}, where $t$ is the time point, $s$ is the seasonal periodicity of the time series and $k$ is the number of sine cosine pairs used with the transformation. 

\begin{equation}
\label{eqn:fourier}
    sin\bigg(\frac{2\pi kt}{s}\bigg), cos\bigg(\frac{2\pi kt}{s}\bigg)
\end{equation}

The number of Fourier terms controls the smoothness of the seasonal pattern. We find the optimal number of Fourier terms to be considered with a particular set of series using a grid-search approach considering the range from 1 to 25.

We use the \textit{auto.arima} function from the \textit{forecast} package \citep{hyndman_2015_forecast} along with the Fourier terms generated by using the \textit{fourier} function to obtain DHR-ARIMA forecasts for a given weekly series.

\paragraph{Recurrent Neural Networks with Long Short-Term Memory Cells}

An RNN is a type of NN that is especially suitable for sequence modelling problems \citep{elman_1990_finding} due to feedback loops in its architecture.
The winning approach of the M4 forecasting competition \citep{smyl_2020_hybrid} in 2018 was based on RNNs, which has given them considerable attention in the forecasting community lately.
We include this model due to heterogeneity considerations. Combining the forecasts of a globally trained RNN with the forecasts obtained from the other locally trained univariate forecasting models incorporates the strengths of both global and local models while mitigating the weaknesses of each other (Section \ref{sec:eval_accuracy}). Extracting seasonal patterns through globality is another benefit of using a global RNN \citep{pablo_2020_principles}.

We use a Long Short-Term Memory (LSTM) with peephole connections \citep{gers_2000_learning}, following the methodology proposed by \citet{hewamalage_2019_recurrent}.
As preprocessing steps, series are normalised by dividing them by their corresponding mean values. Then, the normalised series are transformed to a log scale for variance stabilisation.  

We use two techniques to deal with periodic/seasonality effects in the time series. If the corresponding timestamps are available with each data point in all series and if it is possible to align series across time, then we use Fourier terms \citep{hyndman_2018_fpp} to capture the periodic effects of the weekly time series, where a set of sine and cosine terms are used together with the preprocessed time series when training the RNN model \citep{bandara_2019_msnet}. The \textit{fourier} function in the R package \textit{forecast} \citep{hyndman_2015_forecast} is used for Fourier term calculations. If the corresponding timestamps are not available or it is not possible to align series across time, which is the case,  e.g., in the M4 weekly dataset \citep{makridakis_2018_m4}, then we use a seasonal lag such as 53 to train the RNN model. 
Our RNN model uses a stacked architecture \citep{hewamalage_2019_recurrent, bandara_2019_sales, godahewa_2020_simulation}, with input and output windows for multi-step-ahead forecasting \citep{hewamalage_2019_recurrent}.

Eight hyperparameters need to be chosen during our RNN model training, namely: cell dimension, mini-batch size, maximum number of epochs, epoch size, number of hidden layers, L2-regularisation weight, standard deviation of random normal initialiser and standard deviation of the Gaussian noise. We perform automatic hyperparameter tuning using the Sequential Model based Algorithm Configuration (SMAC) optimisation method \citep{hutter_2011_smac} as described by \citet{hewamalage_2019_recurrent}. The COntinuous COin Betting \citep[COCOB, ][]{orabona_2017_training} algorithm is used as the learning algorithm due to its capability of selecting an optimal learning rate during model training.

\subsubsection{Base Model Pools for Analysis}
\label{sec:weelsel}
We have discussed a general base model pool and a base model pool developed for weekly series. However, following the idea of a parsimonious, heterogeneous base model pool to produce uncorrelated base model forecasts, we further select a subset of most suitable base models that are diverse as well as accurate to make the forecast combination model stronger \citep{Chandra2006}. 

TBATS and DHR-ARIMA are considered as the current state-of-the-art in weekly forecasting, able to capture well the seasonal patterns in such series, so we include them in the selection, as well as the RNN as we have specifically crafted it for weekly data, and it is a global forecasting model able to extract patterns across series. 

We do not consider very simple base models for this model pool, and therefore omit the sna\"ive method. STLM-AR is also omitted as it requires two full periods of data. As discussed in Section~\ref{sec:suitability_weekly}, non-seasonal models are unsuitable for weekly data in principle, but still are often used and lead to good results. Thus, out of the non-seasonal models, we select only one model, namely the Theta method. Theta models the trend in a series as half the slope of a linear regression of the data \citep{hyndman_2003_unmasking}. Thus, it models the trend conservatively and may therefore be better suited for cases where a yearly seasonality in the data is modelled through a trend component. Furthermore, Theta is simple but highly competitive, having been the winning method of the M3 forecasting competition \citep{makridakis_2000_m3}.

Thus, we use the following three base model pools for our analysis.

\begin{description}
  \item [General] This model pool consists of the nine base models used in FFORMA, not focused on forecasting of weekly series in particular: ETS, Auto.ARIMA, TBATS, Theta, na\"ive, sna\"ive, random walk with drift, NNET-AR and STLM-AR. 
  \item [WeekFull] Contains DHR-ARIMA, RNN and seven base models used in FFORMA: ETS, TBATS, Theta, na\"ive, sna\"ive, random walk with drift and STLM-AR.
  \item [WeekSel] Contains the selected four diverse base models that are suitable for weekly forecasting and highly accurate: TBATS, DHR-ARIMA, Theta and a global RNN.
\end{description}

\bigskip

Orthogonally to the choice of base models, we also analyse the effect of applying a logarithmic transformation to the base model forecasts, which is a common preprocessing technique used in the forecasting space to stabilise the variance of forecasts \citep{bandara_2020_clustering, smyl_2020_hybrid}. We apply a logarithm to the base model forecasts before using them with the forecast combination approaches. All time series in our experimental datasets are non-negative. If the base model forecasts contain zeros, then we add a constant $c=1$ before transforming them with the logarithm. 

\subsection{Meta-Learning Architectures}
\label{sec:ensemble_model}

We explore the use of a feature-based meta learning approach and stacking as two meta-learning architectures for weekly forecasting as discussed in the following.

\subsubsection{Feature-based Meta-Learning}
\label{sec:fforma_metalearner}

We refer to the meta-learning architecture used in the FFORMA \citep{pablo_2020_fforma} algorithm as feature-base meta-learning (FBML) in the remainder of the paper. It optimally combines the forecasts from the base models using a set of weights obtained by a gradient boosted tree that has 42 input features. The input features are calculated from the original time series, and they include trend, seasonality, linearity, curvature, and correlation features calculated using the R package \textit{tsfeatures} \citep{hyndman_2019_tsfeatures}.

In the training phase, the algorithm splits each series into a training period and a validation period taken from the end of the series, with a length equal to the forecasting horizon. The features are calculated using the training period. Then, for each base model, the forecasts and the respective errors are calculated for the validation periods of all series. A gradient boosted tree is then trained as a meta-learning model over all series to find a function that maps the features to a set of optimal weights that can be used to combine the base model forecasts in a way that the total loss over the validation periods gets minimised.

In the prediction phase, the features are calculated for the full series and base model forecasts are obtained for the expected forecast horizon. Finally, the base model forecasts are combined using the optimal set of weights provided by the meta-learner given the calculated features.

Some drawbacks of this approach are that it generally requires a large number of long series to learn the sub-model weights. Furthermore, the lengths of the series should be at least two seasonal cycles to calculate the seasonal features, which is oftentimes a difficult requirement for weekly series due to the long seasonal cycle. In particular, with our experimental datasets, most of the seasonal features cannot be calculated for short series with less than two full periods and hence for those, only non-seasonal features are calculated. Simpler meta-learning architectures such as stacking can be considered good alternatives under these circumstances, as discussed in the following section.

\subsubsection{Stacking}
\label{sec:stacking_metalearner}

Stacking \citep{wolpert_1992_stacked} is a widely used meta-learning architecture, and a good option to combine the forecasts of different base model pools. Stacking trains a meta-model by using the forecasts provided by the base models as inputs and it directly outputs the final forecasts.
In the stacking approach, we consider the last sequence of values in a given time series equal to the size of the forecast horizon as its validation part. The base model forecasts are separately computed for this validation part and a meta-learner is trained with them considering the validation part as the true output. Then, the base model forecasts are again separately calculated using the whole time series corresponding with the actual test period. The base model forecasts are provided as the test inputs, and the final forecasts corresponding with the test period are obtained using the previously trained meta-learning model.
Compared with the FBML approach, stacking is capable of learning the sub-model weights with smaller amounts of data. 
If enough data to calculate the features is available, stacking can also be done in a way where both sub-model forecasts and features are used as inputs of the stacked meta-learning model. In the remainder of the paper, we name this strategy StackFeat. In this stacking variant, we use the sub-model forecasts as well as the 42 features used in FBML as inputs during model training.
We also analyse whether training one meta-learner per each horizon or whether training a single global meta-learner across all horizons is the best approach to be used in the weekly forecasting context. When using a single meta-learner, we convert the sub-model forecasts into a vector format before training the meta-learner. In the remainder of the paper, we use PH to denote training multiple meta-learners per each horizon, and S to denote training a single global meta-learner across all horizons.

\subsection{Meta-Learning Algorithms}
\label{sec: algorithms}

For our analysis, we use three algorithms for meta-learning, namely XGBoost \citep{chen_xgboost_2016}, linear regression, and lasso regression \citep{tibshirani_1994_regression}.
XGBoost is a state-of-the-art machine learning algorithm leading usually to very good accuracy, and it is the meta-learner used in FFORMA, which are the reasons for us to use it as a meta-learner in this work. We use the \textit{xgboost} function in the R package \textit{xgboost} \citep{chen_2020_xgboost} to implement the XGBoost meta-learner. The hyperparameters of the XGBoost meta-learner such as $\eta$ and maximum tree depth are tuned using a grid search approach. 
We also consider simpler meta-learning algorithms for our analysis that are suitable with relatively small datasets, as they are common in the forecasting of weekly data.
Linear regression is a simple meta-learner which combines a given set of forecasts. A linear model is fitted using the sub-model forecasts and the forecasts are combined based on their corresponding coefficient values. We use the \textit{glm} function in the R package \textit{glmnet} \citep{friedman_2010_regularization} to implement the linear regression meta-learner.
Model combination generally can be considered more accurate than model selection \citep{KOLASSA2011238}, and model selection can in fact be seen as a special case of model combination. For our analysis, we also consider lasso regression as a meta-learning algorithm which performs both model combination and selection, excluding the worst performing models from the combinations. Lasso regression prevents model overfitting that can occur with the normal linear regression by adding a regularisation term (L1) to the cost function of linear regression. Hence, the coefficients corresponding with the worst performing models shrink towards zero. Prior works of using local lasso regression models for combining forecasts have shown good results \citep{DIEBOLD20191679, Wilms2016LassoBasedFC}. We use the \textit{glmnet} function in the R package \textit{glmnet} \citep{friedman_2010_regularization} to implement the lasso regression model. The regularisation parameter $\lambda$ is tuned as a hyperparameter with 10-fold cross validation.
In the remainder of the paper, we use LR, LinR, and XGB, to denote lasso regression, linear regression, and XGBoost, respectively.

\section{Experimental Framework and Results}
\label{sec:experiments}
In this section, we present our experimental setup and results on seven benchmark datasets for the weekly forecasting models. 

\subsection{Datasets}
\label{sec:datasets}
We use seven publicly available datasets to evaluate the performance of our proposed weekly forecasting models. Two datasets originally contain weekly series and the series in the remaining five datasets are aggregated from lower granularities accordingly to make them weekly. Table \ref{tab:dataset_overview} provides a summary of the datasets, in their weekly aggregated versions used in our work. A brief overview of the datasets is as follows.

\begin{itemize}
    \item M4 Weekly Dataset: The weekly dataset of the M4 forecasting competition \citep{makridakis_2018_m4}.
    \item NN5 Dataset: The dataset of the NN5 forecasting competition contains 111 daily time series of daily cash withdrawals from Automatic Teller Machines (ATM) in the UK \citep{BENTAIEB20127067}. The dataset has missing values which we replace by the median before the temporal aggregation.
    \item Reduced Kaggle Wikipedia Web Traffic Dataset: We use the first 1000 time series from the Kaggle Wikipedia Web Traffic forecasting competition \citep{web_traffic_dataset}. The time series show the number of hits/traffic for a given set of Wikipedia web pages per day. We replace missing values of the dataset by zeros before aggregation.
    \item Ausgrid Energy Dataset: A half-hourly dataset that contains 300 time series representing the general energy consumption of Australian households \citep{ausgrid_dataset}. One series was omitted before aggregation as it has missing values for more than 8 consecutive months.
    \item San Francisco Traffic Dataset: An hourly dataset that contains 862 time series representing the road occupancy rates on San Francisco Bay area freeways from 2015 to 2016 \citep{traffic_dataset, exp_datasets}.
    \item Solar Dataset: A dataset that contains 137 time series representing the solar power production records per every 10 minutes in the state of Alabama in 2006 \citep{solar_dataset, exp_datasets}.
    \item Dominick Dataset: A weekly dataset showing the sales of 28 product categories of Dominick’s Finer Foods, a large American retail chain in the Chicago area \citep{dominicks_dataset}.
\end{itemize}

\begin{table*}
\renewcommand{\arraystretch}{0.6}
\centering
\begin{tabular}{lccccc}
\hline
\textbf{Dataset} & \textbf{No. of Series} & \textbf{Forecast Horizon} &  \textbf{Min. Length} & \textbf{Max. Length} \\
\hline
M4 &  359 & 13 &  80 & 2597 \\
NN5 &  111 & 8 & 105 & 105 \\
Web Traffic & 1000 & 8 & 106 & 106 \\
Ausgrid & 299 & 8 & 148 & 148 \\
Traffic & 862 & 8 & 96 & 96 \\
Solar & 137 & 5 & 44 & 44 \\
Dominick & 28 & 8 & 208 & 391 \\
\hline
\end{tabular}
\caption{Summary of the Used Datasets}
\label{tab:dataset_overview}
\end{table*}

Though many of the datasets have been aggregated from smaller granularities to weekly, we argue that there is no difference between datasets that are ``pure'' weekly datasets and datasets that have been aggregated into weekly datasets from, e.g., daily data. Take as an example sales data in retail. Those will be Point of Sale (POS) data that are time stamped with their exact transaction time throughout the day. As such, daily sales are as much an aggregation as are weekly sales. Taking daily sales and aggregating them further up to weekly sales is no different than aggregating the transaction data directly to a weekly dataset.
 This consideration will hold for any time-stamped transaction data, like ATM withdrawals, ride-share rides, daily webpage hits, and others. This consideration is also valid to electricity and power production data where the weekly electricity usage and power production can be determined by aggregating the corresponding lower granularity data such as hourly/minutely observations.

\subsection{Error Metrics}
\label{sec:error_metric}

We use two error measures that are common in the forecasting research space: sMAPE and  Mean Absolute Scaled Error \citep[MASE, ][]{hyndman_2006_another}, to measure the performance of our models. They are defined in Equations \ref{eqn:smape} and \ref{eqn:mase}, where $M$ is the number of instances in the training set, $S$ is the length of the seasonal cycle of the dataset, $h$ is the forecast horizon, $F_k$ are the generated forecasts and $Y_k$ are the actual values.

\begin{equation}
\label{eqn:smape}
    sMAPE = \frac{100\%}{h}\sum_{k=1}^{h} \frac{|F_{k} - Y_{k}|}{(|Y_{k}| + |F_{k}|)/2} 
\end{equation}

\begin{equation}
\label{eqn:mase} 
    MASE = \frac{\sum_{k=M+1}^{M+h} |F_{k} - Y_{k}|}{\frac{h}{M - S}\sum_{k=S+1}^{M} |Y_{k} - Y_{k - S}|} 
\end{equation}

For datasets containing zeros, namely the Kaggle web traffic and San Francisco traffic datasets in our experiments, we use the variant of the sMAPE proposed by \citet{suilin_2017_kaggle}, which eliminates problems with small values and division by zero by changing $(|Y_{k}| + |F_{k}|)$ in the denominator of the sMAPE as defined in Equation~\ref{eqn:smape} to $max(|Y_{k}| + |F_{k}| + \epsilon, 0.5 + \epsilon)$. We set the parameter $\epsilon$ to its proposed default of 0.1.

The MASE measures the performance of a model compared with the in-sample average performance of a one-step-ahead na\"ive or sna\"ive benchmark. For the M4 dataset, we calculate the MASE using the na\"ive benchmark to compare our results with the original competition results. For the remaining datasets, we choose the  sna\"ive or na\"ive benchmarks depending on the series length. If the dataset is seasonal and has at least one full cycle of data points (52 data points for weekly data), then we calculate the MASE using the sna\"ive benchmark, otherwise using the na\"ive benchmark. 
In particular, the maximum series length of the Solar dataset is 44, so that we use the na\"ive benchmark here. The remaining 5 datasets have series with more than one full cycle of data points, and the sna\"ive benchmark is used. 

To measure the performance of the models across series, we further calculate the mean and median values of sMAPE and MASE across the series. Thus, each model is evaluated using four error metrics: mean sMAPE, median sMAPE, mean MASE, and median MASE, across a dataset.

\subsection{Benchmarks and Variants}
\label{sec:ben_construction}

We use the eleven base models: RNN, DHR-ARIMA, ETS, Auto.ARIMA, TBATS, Theta, na\"ive, sna\"ive, random walk with drift, NNET-AR and STLM-AR as the main benchmarks. 
Furthermore, the averages of the forecasts provided by the base model pools are also considered as benchmarks. We name these benchmarks according to their model pool Average\_General, Average\_WeekFull, and Average\_WeekSel, respectively. 

Based on the base model pools, meta-learning architectures, meta-learning algorithms, number of meta-learners, and the incorporation of features and the logarithmic transformation explained in Section \ref{sec:methodology}, there are 90 different model variants that can be considered for the experiments. To reduce the number of models to run and to make the experiments computationally feasible, as well as to reduce the risk of spurious results, we first perform experiments to determine the best model pool to use, and then focus on this model pool afterwards. Also, as the prediction task is more complex with the FBML approach as it predicts instead of a single forecast a number of weights for the sub-models, the FBML approach is only tested with its proposed meta-learning algorithm, XGB, and we do not consider the simpler lasso regression and linear regression as a meta-learner in this case. We always build one global meta-learner with the FBML approach and logarithmic transformation is not considered, as proposed in the original FFORMA method. 
Thus, out of the 90 possible model variants, 26 model variants are finally compared to identify the best model to forecast weekly series, as further detailed in the results section.

\subsection{Statistical Tests of the Results}
\label{sec:statistical_testing}

We use the non-parametric Friedman rank-sum test to assess the statistical significance of the results provided by different forecasting methods across time series considering a significance level of $\alpha = 0.05$. The methods are ranked on every series of all datasets, namely M4, NN5, Kaggle web traffic, Ausgrid, Traffic, Solar and Dominick, based on their corresponding sMAPE errors. As this procedure gives the same weight to every series in the comparison, datasets with larger amounts of series naturally get a higher weighting. We deem this not a problem in our analysis as the datasets used in our analysis have comparable sizes. 
The best method according to the average rank is chosen as the control method. Furthermore, Hochberg's post-hoc procedure is used to further characterise the statistical differences \citep{garcia_2010_advanced}.

\section{Results and Discussion}
\label{sec:eval_accuracy}

This section details the results in terms of main accuracy and statistical significance, and later also gives some results that provide more insights into the modelling.

\subsection{Main Accuracy Results}

\begin{table}
        \vspace{-9em}
		\centering\fontsize{8}{8}\rm
		\begin{tabular}{rrrrrrrr}
			\toprule
			\cmidrule{2-8}
			& \ M4 & NN5    & Kaggle  & Ausgrid   & Traffic & Solar & Dominick   \\\cmidrule{2-8}
			\addlinespace
			\multicolumn{8}{l}{\bf Mean sMAPE} \\
			\addlinespace
			  Theta & 8.70 & 12.06 & 30.42 & 29.99 & 12.48 & 24.76 & 15.86\\ 
              TBATS & 7.14 & 11.63 & 31.48 & 25.14 & 12.81 & \textbf{\textit{18.93}} & 16.95 \\ 
              DHR-ARIMA & 8.82 & 11.32 & 42.81 & 25.63 & 13.58 & 20.38 & 16.88 \\ 
              RNN & 7.77 & \textbf{\textit{10.38}} & 26.86 & 25.56 & 12.24 & 27.78  & 15.23\\ 
              ETS & 8.86 & 12.22 & 29.84 & 30.01 & 12.44 & 23.40 & 15.86 \\ 
              Naïve &  9.16 & 13.27 & 31.35 & 31.23 & 12.98 & 32.80 & 20.50 \\ 
              Snaïve & 13.94 & 16.48 & 46.96 & 30.17 & 19.42 & 19.94  & 22.82\\ 
              Random Walk &  9.48 & 13.27 & 31.99 & 33.35 & 13.15 & 34.38  & 20.63\\ 
              STLM-AR &  9.21 & 12.45 & 43.71 & 27.05 & 12.73 & 27.37  & 17.30\\ 
              NNET-AR &  9.45 & 14.52 & 39.69 & 26.62 & 14.35 & 27.56  & 17.19\\ 
              Auto.ARIMA &  8.67  & 13.50 & 32.46 & 26.59 & 12.66 & 25.65 & \textbf{\textit{14.71}} \\         
              Average\_General &  6.93  & 10.94 & 30.10 & 23.29 & 12.53 & 25.25 & 16.10 \\ 
               Average\_WeekFull &  \textbf{\textit{6.78}} & 10.81 & 30.02 & 22.99 & 12.27 & 24.71 & 16.08 \\ 
               Average\_WeekSel & 6.84 & \textbf{\textit{10.38}} & \textbf{\textit{29.58}} & \textbf{22.66} & \textbf{\textit{11.67}} & 22.81  & 15.39\\ 
               
               \hline                 
          Stack\_LR\_S\_Log\_WeekFull &  6.69 & 10.32 & \textbf{26.73} & \textbf{\textit{23.02}} & 12.21 & 22.48  & 16.81\\ 
          
Stack\_LinR\_S\_Log\_WeekFull &  6.98 & 10.41 & 27.05 & 23.67 & 12.20 & \textbf{15.67}  & 16.95\\           
          
               Stack\_LR\_S\_Log\_WeekSel & \textbf{6.42} & \textbf{10.16} & \textbf{26.73} & 24.45 & 11.76 & 19.02 & \textbf{\textit{14.61}} \\ 
               Stack\_LinR\_S\_Log\_WeekSel &  6.71 & 10.17 & 26.93 & 24.10 & \textbf{\textit{11.63}} & 19.18  & 15.57\\              

               \hline           
              
                     Stack\_LR\_PH\_Log\_WeekSel & 6.93 & 12.04 & \textbf{\textit{26.82}} & 25.85 & 11.26 & 20.57  & 17.10\\ 
                     Stack\_LinR\_PH\_Log\_WeekSel &  6.94 & 12.13 & 27.06 & 24.94 & \textbf{11.05} & 21.49  & 17.61\\ 
                      StackFeat\_LR\_S\_WeekSel & 7.07 & \textbf{\textit{10.55}} & 27.19 & \textbf{\textit{22.87}} &    11.99 & 18.92  & 24.98\\ 
                              
             StackFeat\_XGB\_S\_WeekSel &  8.47 & 13.32 & 34.38 & 30.85 & 13.74 & \textbf{\textit{16.92}}  & 30.20\\ 
                                    
              StackFeat\_LR\_S\_Log\_WeekSel & \textbf{\textit{6.87}} & 10.65 & 27.27 & 25.34 &  11.96 &  18.77 & \textbf{\textit{16.98}} \\                                
              
              \hline 
              
              FBML\_XGB\_S\_General & 7.91 & 13.34 & 34.30 & 26.45 &    12.41 &  24.98 &  18.19 \\ 
              FBML\_XGB\_S\_WeekSel & \textbf{\textit{6.96}} & \textbf{\textit{10.66}} & \textbf{\textit{32.43}} & \textbf{\textit{24.46}} &    \textbf{\textit{12.01}} &    \textbf{\textit{23.50}}  & \textbf{14.36} \\        
            \bottomrule
			\addlinespace

			\multicolumn{8}{l}{\bf Median sMAPE} \\
			\addlinespace
		      Theta & 5.40 & 10.97 & 26.20 & 24.31 &  9.72 & 24.90  &  13.31\\ 
              TBATS & 4.94 & 11.15 & 27.28 & 17.83 & 10.00 & \textbf{\textit{18.02}}  &  14.51\\ 
              DHR-ARIMA & 5.12 & 11.07 & 34.34 & 18.98 &  \textbf{\textit{8.39}} & 19.60  &  14.04\\ 
              RNN & 5.09 &  9.94 & \textbf{\textit{24.25}} & 19.15 &  9.60 & 27.69  &  \textbf{\textit{12.55}}\\ 
              ETS & 5.15 & 10.85 & 25.91 & 24.82 & 9.77 & 24.51 &  13.76 \\
              Naïve & 5.18 & 10.89 & 27.49 & 26.28 & 10.20 & 32.07  &  15.67\\ 
              Snaïve & 8.36 & 15.40 & 40.01 & 21.96 & 14.83 & 20.38  &  20.69\\ 
              Random Walk & 5.16 & 10.99 & 27.46 & 27.87 & 10.30 & 33.90  &  15.62\\ 
              STLM-AR & 6.46 & 11.16 & 37.99 & 21.39 & 10.27 & 27.28  &  15.06\\ 
              NNET-AR & 5.05 & 12.83 & 31.70 & 19.98 &  9.68 & 27.19  &  14.39\\ 
              Auto.ARIMA & 5.25  & 12.22 & 26.48 & 20.06 & 10.01 & 25.68  &  12.71\\ 
              
              Average\_General & 4.73  & 10.02 & 26.31 & 17.10 &  9.72 & 25.13  &  14.10\\
               Average\_WeekFull & 4.84 &  \textbf{\textit{9.78}} & 26.32 & 17.27 &  9.54 & 24.79 &  14.46 \\ 
               Average\_WeekSel & \textbf{\textit{4.51}} &  9.86 & 25.61 & \textbf{16.13} &  8.75 & 22.54  &  14.14\\ 
              \hline
              Stack\_XGB\_S\_Log\_General & 6.38  & 13.77 & 30.77 & 23.20 & 11.25 & 21.62  &  \textbf{12.44}\\ 
              Stack\_LR\_S\_Log\_WeekFull & 4.62 &  9.83 & 24.30 & \textbf{\textit{16.74}} &  9.60 & 22.00  &  13.88\\               
              Stack\_LR\_S\_Log\_WeekSel & \textbf{4.08} &  9.56 & \textbf{24.13} & 18.26 &  9.03 & 19.03  &  13.10\\ 
              Stack\_LinR\_S\_Log\_WeekSel & 4.40 &  \textbf{9.49} & 24.38 & 17.04 &  \textbf{\textit{8.83}} & 19.46 &  13.43 \\
              Stack\_XGB\_S\_Log\_WeekSel & 6.15 & 12.48 & 30.32 & 26.05 & 11.62 & \textbf{15.99}  &  19.02\\ 
              \hline
              
               Stack\_LR\_PH\_Log\_WeekSel & 4.60 & 11.52 & \textbf{\textit{24.42}} & 19.37 &  8.31 & 19.66  &  14.73\\ 
               Stack\_LinR\_PH\_Log\_WeekSel & 4.58 & 11.63 & 24.63 & 18.41 &  \textbf{8.02} & 21.24  &  \textbf{\textit{13.83}}\\ 
              StackFeat\_LR\_S\_WeekSel & \textbf{\textit{4.28}} & \textbf{\textit{9.70}} & 24.69 & \textbf{\textit{16.62}} &    9.16 &  18.91  &  15.41\\              
              StackFeat\_XGB\_S\_Log\_WeekSel & 6.55 & 12.49 & 28.83 & 23.06 & 10.58 & \textbf{\textit{16.13}}  &  16.28\\ 
              \hline
              
              FBML\_XGB\_S\_General & \textbf{\textit{4.52}} & 11.92 & 27.93 & 19.78 &   9.63 & 25.01  &  13.23\\ 
              FBML\_XGB\_S\_WeekSel & 4.69 & \textbf{\textit{10.10}} & \textbf{\textit{26.46}} & \textbf{\textit{17.01}} &   \textbf{\textit{9.50}} & \textbf{\textit{23.54}}  &  \textbf{\textit{12.61}}\\                         
            \bottomrule
			\end{tabular}
		\caption{sMAPE results for methods that perform best overall and per category across all experimental datasets. The results of all model variants across all error metrics are available in the Online Appendix. The best performing models in each group are italicized and the overall best performing models are highlighted in boldface. 
		}\label{tab:all_results}
\end{table}

\begin{table}
        \vspace{-9em}
		\centering\fontsize{8}{8}\rm
		\begin{tabular}{rrrrrrrr}
			\toprule
			\cmidrule{2-8}
			& \ M4 & NN5    & Kaggle  & Ausgrid   & Traffic & Solar & Dominick   \\\cmidrule{2-8}
			\addlinespace
			\multicolumn{8}{l}{\bf Mean MASE} \\
			\addlinespace
			  Theta & 2.734 & 0.898 & 0.688 & 1.206 & 1.122 & 1.224  &  0.877\\ 
              TBATS & 2.249 & 0.864 & 0.692 & 1.045 & 1.149 & \textbf{\textit{0.910}}  &  0.936\\ 
              DHR-ARIMA & 2.418 & 0.854 & 0.834 & 1.068 & 1.023 & 0.991  &  0.926\\ 
              RNN & 2.482 & \textbf{\textit{0.768}} & \textbf{\textit{0.608}} & 1.042 & 1.128 & 1.422  &  \textbf{\textit{0.811}}\\ 
              ETS & 2.334 & 0.902 & 0.685 & 1.209 & 1.122 & 1.158  &  0.870\\
              Naïve & 2.777 & 0.974 & 0.738 & 1.258 & 1.178 & 1.735  &  1.253\\ 
              Snaïve & 9.736 & 1.160 & 1.016 & 1.229 & 1.581 & 0.942  &  1.197\\ 
              Random Walk & 2.682 & 0.977 & 0.751 & 1.316 & 1.191 & 1.842  &  1.262\\ 
              STLM-AR & 3.519 & 0.915 & 0.884 & 1.150 & 1.135 & 1.382  &  0.958\\ 
              NNET-AR & 3.015 & 1.099 & 1.236 & 1.090 & 1.152 & 1.410  &  0.988\\ 
              Auto.ARIMA &  2.377  & 1.000 & 0.696 & 1.134 & 1.131 & 1.285  &  0.829\\              
              Average\_General & 2.742 & 0.812 & 0.704 & 0.952 & 1.110 & 1.257  &  0.893\\
              Average\_WeekFull & 2.698 & 0.802 & 0.663 & 0.941 & 1.089 & 1.227  &  0.893\\ 
              Average\_WeekSel & \textbf{\textit{2.142}} & 0.777 & 0.653 & \textbf{0.930} & \textbf{\textit{1.020}} & 1.119  &  0.852\\ 
              \hline            
              Stack\_LR\_S\_Log\_WeekFull & 2.227 & 0.772 & 0.607 & \textbf{\textit{0.945}} & 1.112 & 1.098  &  0.932\\ 
              Stack\_LinR\_S\_Log\_WeekFull & 2.208 & 0.774 & 0.612 & 0.965 & 1.117 & \textbf{0.732}  &  0.921\\ 
              
              Stack\_LR\_S\_Log\_WeekSel & \textbf{2.112} & \textbf{0.761} & \textbf{0.606} & 1.004 & 1.053 & 0.905  &  \textbf{\textit{0.802}}\\
              Stack\_LinR\_S\_Log\_WeekSel & 2.281 & 0.763 & 0.610 & 0.982 & \textbf{\textit{1.039}} & 0.926  &  0.857\\ 
              
              \hline
                     
              Stack\_LR\_PH\_Log\_WeekSel & 2.599 & 0.909 & \textbf{\textit{0.608}} & 1.074 & 1.024 & 1.013  &  0.957\\   
              Stack\_LinR\_PH\_Log\_WeekSel & 2.637 & 0.918 & 0.612 & 1.028 & \textbf{1.004} & 0.989  &  0.958\\ 
              
              StackFeat\_LR\_PH\_Log\_WeekSel & 2.434 & 0.941 & 0.616 & 1.080 &    1.064 &    1.126  &  1.041\\ 
               
              StackFeat\_LR\_S\_WeekSel & 2.340 & \textbf{\textit{0.784}} & 0.617 & \textbf{\textit{0.948}} &    1.111 &    0.901  &  1.181\\ 
              StackFeat\_LR\_S\_Log\_WeekSel & \textbf{\textit{2.259}} & 0.794 & 0.617 & 1.051 &    1.108 &  0.895  &  \textbf{\textit{0.941}}\\ 
             StackFeat\_XGB\_S\_Log\_WeekSel & 4.002 & 0.954 & 0.728 & 1.195 & 1.264 & \textbf{\textit{0.755}}  &  1.316\\ 
               
              \hline
              
              FBML\_XGB\_S\_General & \textbf{\textit{2.156}} & 0.944 & 0.723 & 1.062 &    1.683 &    1.249  &  1.097\\ 
              FBML\_XGB\_S\_WeekSel & 2.194 & \textbf{\textit{0.796}} & \textbf{\textit{0.659}} & \textbf{\textit{0.994}} &    \textbf{\textit{1.541}} &    \textbf{\textit{1.159}}  &  \textbf{0.789}\\           
         
            \bottomrule
			\addlinespace
			\multicolumn{8}{l}{\bf Median MASE} \\
			\addlinespace
		      Theta & 1.969 & 0.805 & 0.549 & 0.981 & 0.983 & 1.241  &  0.843\\ 
              TBATS & \textbf{1.414} & 0.874 & 0.553 & 0.848 & 0.997 & \textbf{\textit{0.892}}  &  0.795\\ 
              DHR-ARIMA & 1.640 & 0.826 & 0.686 & 0.903 & \textbf{0.793} & 0.967  &  0.778\\ 
              RNN & 1.582 & \textbf{0.680} & \textbf{0.483} & 0.867 & 0.924 & 1.426  &  \textbf{\textit{0.697}}\\ 
              ETS & 1.801 & 0.765 & 0.534 & 0.992 & 0.978 & 1.227  &  0.856\\
              Naïve & 1.938 & 0.776 & 0.556 & 1.046 & 1.023 & 1.707  &  1.022\\ 
              Snaïve & 3.099 & 1.115 & 0.838 & 1.026 & 1.442 & 0.974  &  1.156\\ 
              Random Walk & 1.864 & 0.822 & 0.557 & 1.076 & 1.035 & 1.824  &  1.023\\ 
              STLM-AR & 2.053 & 0.867 & 0.738 & 0.986 & 0.953 & 1.400  &  0.857\\ 
              NNET-AR & 1.963 & 0.976 & 0.678 & 0.928 & 0.924 & 1.350  &  0.868\\ 
              Auto.ARIMA & 1.660 & 0.936 & 0.547 & 0.905 & 0.997 & 1.289  &  0.843\\              
              Average\_General & 1.680 & 0.749 & 0.537 & 0.753 & 0.956 & 1.269  &  0.815\\ 
              Average\_WeekFull & 1.674 & 0.703 & 0.531 & \textbf{0.724} & 0.930 & 1.234  &  0.810\\ 
              Average\_WeekSel & 1.514 & 0.730 & 0.516 & 0.730 & 0.850 & 1.114  &  0.704\\ 
              
              \hline           
              Stack\_LR\_S\_Log\_WeekFull & 1.483 & \textbf{\textit{0.690}} & \textbf{\textit{0.485}} & 0.756 & 0.953 & 1.115  &  0.808\\ 
              Stack\_LinR\_S\_Log\_WeekFull & \textbf{\textit{1.423}} & 0.726 & 0.488 & \textbf{\textit{0.752}} & 0.942 & \textbf{0.749}  &  0.824\\ 
              
              Stack\_LR\_S\_Log\_WeekSel & 1.512 & 0.691 & \textbf{\textit{0.485}} & 0.799 & 0.879 & 0.936  &  \textbf{\textit{0.712}}\\
              Stack\_LinR\_S\_Log\_WeekSel & 1.497 & 0.694 & 0.487 & 0.777 & \textbf{\textit{0.848}} & 0.960  &  0.770\\ 
              
              \hline
             
              Stack\_LR\_PH\_Log\_WeekSel & 1.836 & 0.841 & \textbf{\textit{0.484}} & 0.891 & 0.816 & 0.980  &  0.909\\ 
              Stack\_LinR\_PH\_Log\_WeekSel & 1.793 & 0.846 & 0.485 & 0.843 & \textbf{\textit{0.798}} & 0.986  &  \textbf{\textit{0.815}}\\ 
             
              StackFeat\_LR\_S\_WeekSel & 1.607 & 0.753 & 0.501 & \textbf{\textit{0.726}} &    0.891 &    0.929  &  1.026\\ 
              StackFeat\_LR\_S\_Log\_WeekSel & \textbf{\textit{1.492}} & \textbf{\textit{0.739}} & 0.496 & 0.819 &    0.888 &    0.922  &  0.886\\              
             StackFeat\_XGB\_S\_Log\_WeekSel & 2.054 & 0.918 & 0.582 & 0.979 & 1.018 & \textbf{\textit{0.755}}  &  0.959\\ 
              
              \hline
              
              FBML\_XGB\_S\_General & 1.569 & 0.899 & 0.553 & 0.874 &    1.372 &    1.272  &  0.823\\ 
              FBML\_XGB\_S\_WeekSel & \textbf{\textit{1.560}} & \textbf{\textit{0.696}} & \textbf{\textit{0.528}} & \textbf{\textit{0.775}} &    \textbf{\textit{1.125}} &    \textbf{\textit{1.155}}  &  \textbf{0.644}\\   
              \bottomrule
		\end{tabular}
		\caption{MASE results for methods that perform best overall and per category across all experimental datasets. The best performing models in each group are italicized and the overall best performing models are highlighted in boldface.}\label{tab:all_results_mase}
\end{table}

Tables \ref{tab:all_results} and \ref{tab:all_results_mase} present the results of a selected set of model variants and benchmarks across all experimental datasets for mean/median sMAPE and mean/median MASE, respectively. The results of all model variants across all error metrics are available in an Online Appendix\footnote{The Online Appendix is available at \url{https://drive.google.com/file/d/1XrZUlnvLUit6MQkqt3D_d4MuC9nX8Jo9/view?usp=sharing}.}.

\subsubsection{Naming of Model Combinations}
\label{sec:naming}

The model combinations in Tables \ref{tab:all_results} and \ref{tab:all_results_mase}, and Tables A.1 and A.2 in the Appendix are named in a way that they explain the corresponding used base model pool, meta-learning architecture, meta-learning algorithm, number of meta-learners and whether log transformation is applied as a preprocessing technique. In particular the model combination names have the following structure.

\thickmuskip=0mu
$<architecture>\_<algorithm>\_<number-mls>\_<log>\_<model-pool>$

\bigskip
The meta-learning architecture (Section \ref{sec:ensemble_model}) is one of FBML, Stack or StackFeat (stacking with features). The meta-learning algorithm (Section \ref{sec: algorithms}) is one of lasso regression (LR), linear regression (LinR) or XGBoost (XGB). The number of meta-learners (Section \ref{sec:stacking_metalearner}) can be either one meta-learner per each horizon (PH) or single meta-learner (S). The base model pool (Section \ref{sec:base_models}) is one of General, WeekFull or WeekSel.

\subsubsection{Grouping of Sub-Experiments}
\label{sec:grouping}
The models in Tables \ref{tab:all_results} and \ref{tab:all_results_mase}, and Tables A.1 and A.2 in the Appendix are grouped based on the sub-experiments. The results of the best-performing models in each group are italicized, and the overall best performing models across the datasets are highlighted in boldface. 
We see that in the first group which contains the benchmark models, the Average\_WeekSel method performs best for all datasets except the M4, Solar and Dominick datasets. On the M4 dataset, Average\_WeekFull performs the best in the first group,  on the Solar dataset, TBATS and Sna\"ive perform better and on the Dominick dataset, Auto.ARIMA and RNN perform better.

These results already give an indication that the WeekSel model pool may be the most suitable model pool for weekly time series forecasting. In the next experiment, we further focus on the question which base model pool is the most suitable one.

To make the experiments more tractable, we fix the meta-learning architecture in the following experiment, under the assumption that the meta-learning architecture will have little influence on which base-model pool works well, compared with the choice of meta-learning algorithms which will have a larger influence. E.g., the lasso regression meta-learning algorithm has the capability to drop sub-models when their contribution is negligible to the final forecasts, so that we assume that some meta-learning algorithms can work better with larger model pools than others.
We consider a model configuration that uses the Stack architecture and trains a single global meta-learner across all horizons (S).
By training a single meta-learner, it is the simplest architecture among all architectures considered in our work. Furthermore, we use the log transformed sub-model forecasts, as from preliminary experiments not reported here, we observe that training the model with log transformed forecasts provides better results for the experimental datasets compared with the models that use forecasts in the original scale.
This simple model configuration is tested with all 3 base model pools (General, WeekFull, WeekSel) combined with three meta-learning algorithms (LR, LinR, XGB), making 9 different model combinations. 

As seen in the second group of models in Tables A.1 and A.2 in the Appendix, in general the variants provide better results with the WeekSel model pool across the experimental datasets and meta-learning algorithms. Hence, we limit the subsequent experiments to this model pool, discarding the other two model pools. 

In the third experiment, we further analyse the usage of incorporating features during the training of stacking models (StackFeat), the effect of applying log transformation for sub-model forecasts and training multiple meta-learners one per each horizon (PH) in contrast to training a single global meta-learner. All possible modelling combinations of StackFeat are executed with the best base model pool, WeekSel as shown in Tables A.1 and A.2 in the Appendix. The variants that use lasso regression generally show better performance in this group. The StackFeat architecture also shows a better performance when it builds a single meta-learner across all horizons. The usage of log transformed forecasts shows a mixed performance with the StackFeat architecture based variants. 

In Tables \ref{tab:all_results} and \ref{tab:all_results_mase}, and Tables A.1 and A.2 in the Appendix, in the last group of models, FBML\_XGB\_S\_General refers to the original FFORMA approach. As explained in Section \ref{sec:ben_construction}, the FBML approach is only tested under its proposed settings by only building one global meta-learner with the XGBoost meta-learning algorithm without considering logarithmic transformation. The FBML approach is also tested with the best model pool, WeekSel, and we see that using this model pool leads to consistently improved results.

\subsubsection{Best-performing Model Variant: Stack\_LR\_S\_Log\_WeekSel}
\label{sec:best_variant}

\begin{sloppypar}
Based on all considered variants shown in Tables A.1 and A.2 in the Appendix, Stack\_LR\_S\_Log\_WeekSel overall provides the best forecasts. In particular, we see that on mean sMAPE, Stack\_LR\_S\_Log\_WeekSel outperforms all base models for all experimental datasets except for the Solar dataset. Furthermore, it outperforms all benchmarks and variants on the M4, NN5 and Kaggle web traffic datasets. For the M4 weekly dataset, Stack\_LR\_S\_Log\_WeekSel produces a mean sMAPE of 6.42, which places it at the first position in the original results of the M4 weekly competition \citep{m4_results_2018}, with an sMAPE result of 6.58 of the original top solution.
\end{sloppypar}

Overall, Stack\_LR\_S\_Log\_WeekSel demonstrates a better performance on median sMAPE compared with the other benchmark models and variants where it is the best performing model on M4 and Kaggle web traffic datasets. 

On mean MASE, Stack\_LR\_S\_Log\_WeekSel outperforms the individual base models for all experimental datasets except the Traffic dataset. Furthermore, it is the best-performing model on the M4, NN5, and Kaggle web traffic datasets. For the M4 weekly dataset, Stack\_LR\_S\_Log\_WeekSel produces a mean MASE of 2.112, which places it at the third position in the original results of the M4 weekly competition. In the original results, FFORMA was at the second position with a mean MASE of 2.108. A difference between the original participating solution of FFORMA in the M4 competition and FBML\_XGB\_S\_General in our experiments is that those authors use the full M4 dataset containing 100,000 series for model training. But in our case, as our focus is on weekly series, we only use the 359 weekly series in the M4 dataset for FBML\_XGB\_S\_General model training which produces a mean MASE of 2.156. Thus, FBML\_XGB\_S\_General in this configuration does not outperform Stack\_LR\_S\_Log\_WeekSel. 

On median MASE, the performance is slightly different. The best method on median MASE is a single base model for all datasets, except the Ausgrid, Solar and Dominick datasets, which also shows the suitability of using a pool of baseline models to forecast weekly series.

However, across the mean sMAPE and mean MASE results, the best performing model on each dataset is the same where across three datasets Stack\_LR\_S\_Log\_WeekSel performs the best. Furthermore, the best performing models across different model groups are also the same across mean sMAPE and mean MASE results on all datasets except for five cases: M4 dataset best performing models on the first and last groups, Kaggle web traffic dataset best performing model on the first group, Solar dataset best performing model on the third group and Dominick dataset best performing model on the first group. Thus, overall, the best performing models across sMAPE and MASE are very similar and they only have minor differences.

\subsubsection{Reasons for the Better Performance of Stack\_LR\_S\_Log\_WeekSel}
\label{sec:best_meta_algorithm}

\begin{sloppypar}
Stack\_LR\_S\_Log\_WeekSel also shows a better performance when comparing with the variants shown in Tables A.1 and A.2 in the Appendix that use only the forecasts of the sub-models in WeekSel with different meta-learning algorithms during training. Stack\_LinR\_S\_Log\_WeekSel only outperforms Stack\_LR\_S\_Log\_WeekSel on Ausgrid and Traffic datasets across all error metrics. Stack\_LinR\_S\_Log\_WeekSel also outperforms Stack\_LR\_S\_Log\_WeekSel on M4 and NN5 weekly datasets only across median MASE and median sMAPE respectively. However, overall Stack\_LR\_S\_Log\_WeekSel demonstrates a better performance than Stack\_LinR\_S\_Log\_WeekSel on five datasets. Furthermore, Stack\_LR\_S\_Log\_WeekSel outperforms Stack\_XGB\_S\_Log\_WeekSel on all datasets across all error metrics except for the Solar dataset. Hence, Tables A.1 and A.2 in the Appendix clearly show that in our case, lasso regression is a better option to be used as a meta-learning algorithm to combine sub-model forecasts than other regression techniques such as linear regression and XGBoost, due to its capability of sub-model selection (Section \ref{sec: algorithms}).
\end{sloppypar}

\begin{sloppypar}
Although lasso regression is a robust algorithm that has the capability to select models, when using more suitable sub-models, the final forecasts get more accurate (Section \ref{sec:base_models}). Hence, even though the WeekFull model pool contains the four base models in WeekSel, overall, Stack\_LR\_S\_Log\_WeekFull provides less accurate forecasts compared with Stack\_LR\_S\_Log\_WeekSel. The general model pool contains the original FFORMA sub-models. Overall, Stack\_LR\_S\_Log\_General provides less accurate forecasts across all datasets compared with both Stack\_LR\_S\_Log\_WeekSel and Stack\_LR\_S\_Log\_WeekFull showing that including more suitable base models can considerably improve forecast accuracy for weekly series. 
However, Stack\_LR\_S\_Log\_General outperforms FBML\_XGB\_S\_General across all datasets on all error metrics except the mean and median sMAPE of the Traffic dataset and median sMAPE of the Dominick dataset. It further shows that using a lasso regression based stacking approach is a better option to combine sub-model forecasts in our considered weekly forecasting datasets than the FBML approach (Section \ref{sec:stacking_metalearner}). Compared with FBML\_XGB\_S\_General, Stack\_LR\_S\_Log\_WeekSel uses more suitable base models and a meta-learning algorithm to produce final forecasts. Hence, Stack\_LR\_S\_Log\_WeekSel outperforms FBML\_XGB\_S\_General across all datasets on all error metrics. Furthermore, a linear model such as lasso regression can be more suitable as a forecast combination method than FBML\_XGB\_S\_General, if there is only a small amount of data available for model training (Section \ref{sec: algorithms}). Stack\_LR\_S\_Log\_WeekSel also contains the strengths of both global and local models. As a result, it demonstrates a better performance than FBML\_XGB\_S\_General and the other benchmark models. 
\end{sloppypar}

FBML\_XGB\_S\_WeekSel has performed better compared with FBML\_XGB\_S\_General across all experimental datasets on all error metrics except the median sMAPE and mean MASE of the M4 weekly dataset. This further indicates that the base models in WeekSel are more suitable to forecast weekly data than the sub-forecasting models in the General model pool (Section \ref{sec:weelsel}). However, on the NN5, Kaggle web traffic, Traffic and Solar datasets, FBML\_XGB\_S\_WeekSel has not been able to outperform some of the base models, underlining the need to explore other combination approaches. Stack\_LR\_S\_Log\_WeekSel outperforms FBML\_XGB\_S\_WeekSel across all datasets on all error metrics except for 7 cases: median sMAPE, mean MASE and median MASE of Ausgrid dataset, and all error metrics on the Dominick dataset. 

\subsubsection{Performance of Simple Averaging and Stack\_LR\_S\_Log\_WeekSel}
\label{sec:best_method_vs_avg}

Generally, averaging the base model forecasts is considered as an efficient and accurate ensembling technique in the forecasting research space. Simple averaging has outperformed FBML\_XGB\_S\_WeekSel in many cases even though the FBML approach starts optimising weights considering the simple average.  Major differences between the validation and test sets and small sample sizes can be the reasons for this phenomenon. But simple averaging has only outperformed Stack\_LR\_S\_Log\_WeekSel on the Ausgrid and Traffic datasets on all error metrics. Averaging does not select the best forecasting models when producing the final forecasts. Furthermore, it assigns equal weights to all base models and hence, if a subset of models provides poor forecasts, then the final result may considerably be affected by it. This can be a problem especially if there are models that may be clearly unsuitable, as in our case where we use non-seasonal models to forecast seasonal time series. Stack\_LR\_S\_Log\_WeekSel addresses both of the above-mentioned issues of simple averaging and hence, overall, it has produced more accurate forecasts than the averaging benchmark. The conclusions are the same with other base model pools as well, where overall, both Average\_WeekSel and Stack\_LR\_S\_Log\_WeekSel outperform Average\_General and Average\_WeekFull on all datasets across all error metrics. 

\subsubsection{Time Series Features as Additional Inputs}
\label{sec:feature_effect}

Next, we analyse whether including time series features as inputs can improve the forecasting accuracy. The experiments with features are only conducted with the WeekSel model pool as it is the best base model pool to be used in the weekly forecasting context compared to the other considered two base model pools. The performance of lasso regression models that only take base model forecasts as inputs, namely Stack\_LR\_S\_Log\_WeekSel and Stack\_LR\_PH\_Log\_WeekSel, is generally better than the performance of lasso regression models that take both forecasts and features as inputs: StackFeat\_LR\_S\_Log\_WeekSel, StackFeat\_LR\_S\_WeekSel, StackFeat\_LR\_PH\_Log\_WeekSel and StackFeat\_LR\_PH\_WeekSel. The observations are also the same across the variants that use the other two meta-learning algorithms, linear regression and XGBoost. Hence, it is clear that including features as inputs in our case does not increase the accuracy. Furthermore, for datasets with short time series, incorporating features to train models can pose additional problems as seasonality is not taken into account during feature calculation.

\subsubsection{Number of Meta-Learners}
\label{sec:meta_num}

We also observe that in our case, training a single meta-learner across all horizons provides better results compared to the models which train separate meta-learners per each horizon among the variants that use the best model pool, WeekSel, across all three meta-learning algorithms. Furthermore, our experiments show that transforming the base model forecasts with a logarithm is beneficial, when training a meta-learning model in a weekly forecasting context. 

\bigskip

In summary, Stack\_LR\_S\_Log\_WeekSel shows a better performance compared with all other benchmarks and variants across all datasets. In particular, it provides the best forecasts for the M4 weekly dataset with a mean sMAPE of 6.42 which is better than the performance of the winning method of the M4 weekly competition \citep{goodrich_2020_forecast_pro} with a mean sMAPE of 6.58. The M4 weekly winning approach is not an automated forecasting framework. It uses more than 15 sub-model forecasts where domain knowledge and manual inspection are also required to obtain the final forecasts. Compared with the M4 weekly winning approach, Stack\_LR\_S\_Log\_WeekSel uses only four sub-models with a fully-automated forecast combination approach without using any specific domain knowledge. Hence, we propose Stack\_LR\_S\_Log\_WeekSel as an accurate and a fully-automated strong ensemble model in the weekly time series forecasting space.

\subsection{Statistical Testing Results}

Table \ref{tab:statistical_testing} shows the results of the statistical testing evaluation, namely the adjusted $p$-values calculated from the Friedman test with Hochberg’s post-hoc procedure considering a significance level of $\alpha = 0.05$ \citep{garcia_2010_advanced}. The overall $p$-value of the Friedman rank sum test is less than $10^{-30}$ which is highly significant. Stack\_LR\_S\_Log\_WeekSel performs the best on ranking over sMAPE per each series of all experimental datasets and hence, it is used as the control method as mentioned in the first row.
A horizontal line is used to separate the models that perform significantly worse than Stack\_LR\_S\_Log\_WeekSel. All individual models, ensemble benchmarks and variants are significantly worse than the control method except Stack\_LinR\_S\_Log\_WeekSel as they report $p_{Hoch}$ values less than $\alpha$. 

\begin{table}
\renewcommand{\arraystretch}{0.6}
\centering
\begin{tabular}{lc}
  \hline
  Model & $p_{Hoch}$ \\ 
  \hline
   Stack\_LR\_S\_Log\_WeekSel & - \\ 
   Stack\_LinR\_S\_Log\_WeekSel & $0.42$ \\
   \hline
   Stack\_LR\_PH\_Log\_WeekSel & $0.01$ \\
   Stack\_LinR\_PH\_Log\_WeekSel & $1.74 \times 10^{-3}$ \\
   Average\_WeekSel & $1.26 \times 10^{-6}$ \\
   StackFeat\_LR\_S\_WeekSel & $1.22 \times 10^{-6}$ \\
   Stack\_LR\_S\_Log\_WeekFull & $7.48 \times 10^{-7}$ \\
   Stack\_LinR\_S\_Log\_WeekFull & $4.51 \times 10^{-8}$ \\
   StackFeat\_LR\_S\_Log\_WeekSel & $3.95 \times 10^{-9}$ \\
   StackFeat\_LR\_PH\_Log\_WeekSel & $5.98 \times 10^{-16}$ \\
   RNN & $<10^{-30}$ \\
   StackFeat\_LinR\_S\_Log\_WeekSel & $<10^{-30}$ \\
   Stack\_LinR\_S\_Log\_General & $<10^{-30}$ \\
   FBML\_XGB\_S\_WeekSel & $<10^{-30}$ \\
   Average\_WeekFull & $<10^{-30}$ \\
   Stack\_LR\_S\_Log\_General & $<10^{-30}$ \\ 
   Average\_General & $<10^{-30}$ \\
   TBATS & $<10^{-30}$ \\
   ETS & $<10^{-30}$\\
   StackFeat\_LinR\_PH\_Log\_WeekSel & $<10^{-30}$ \\ 
   StackFeat\_LinR\_S\_WeekSel& $<10^{-30}$ \\
   FBML\_XGB\_S\_General & $<10^{-30}$ \\
   Theta & $<10^{-30}$ \\
   Auto.ARIMA & $<10^{-30}$ \\
   StackFeat\_LR\_PH\_WeekSel & $<10^{-30}$ \\
   DHR-ARIMA & $<10^{-30}$ \\
   StackFeat\_XGB\_S\_Log\_WeekSel & $<10^{-30}$ \\
   NNET-AR & $<10^{-30}$ \\
   Na\"ive & $<10^{-30}$ \\
   StackFeat\_XGB\_S\_WeekSel & $<10^{-30}$ \\
   StackFeat\_LinR\_PH\_WeekSel & $<10^{-30}$ \\
   Random Walk & $<10^{-30}$ \\
   Stack\_XGB\_S\_Log\_WeekFull & $<10^{-30}$ \\
   Stack\_XGB\_S\_Log\_WeekSel & $<10^{-30}$ \\
   StackFeat\_XGB\_PH\_WeekSel & $<10^{-30}$ \\
   Stack\_XGB\_S\_Log\_General & $<10^{-30}$ \\
   Stack\_XGB\_PH\_Log\_WeekSel & $<10^{-30}$ \\
   StackFeat\_XGB\_PH\_Log\_WeekSel & $<10^{-30}$ \\
   STLM-AR & $<10^{-30}$ \\
   Sna\"ive & $<10^{-30}$ \\
  \hline
\end{tabular}
\caption{Results of Statistical Testing} 
\label{tab:statistical_testing}
\end{table}

\begin{sloppypar}
Table \ref{tab:win_loss} shows a comparison of our proposed method, Stack\_LR\_S\_Log\_WeekSel with the M4 weekly winning method, and all other considered benchmarks and variants using the M4 weekly dataset. For each method, we calculate across how many series Stack\_LR\_S\_Log\_WeekSel wins, loses and ties in terms of the sMAPE of each series in the M4 weekly dataset. As shown in Table \ref{tab:win_loss}, our proposed method obtains more wins than losses over all other methods including the winning method of the M4 weekly competition. This further shows that our proposed method is a more accurate weekly forecasting model compared to all other considered benchmarks and variants.
\end{sloppypar}

\begin{table}
\renewcommand{\arraystretch}{0.6}
    \centering
    \begin{tabular}{lrrr}
    \hline
        Model & Win & Loss & Tie \\ \hline
        M4 Weekly Winning Method & 195 & 164 & 0 \\ 
        Stack\_LinR\_S\_Log\_WeekSel & 196 & 162 & 1 \\ 
        StackFeat\_LR\_S\_Log\_WeekSel & 199 & 158 & 2 \\ 
        Stack\_LR\_S\_Log\_WeekFull & 206 & 152 & 1 \\ 
        FBML\_XGB\_S\_General & 206 & 149 & 4 \\ 
        Average\_General & 211 & 143 & 5 \\ 
        Stack\_LinR\_S\_Log\_WeekFull & 212 & 144 & 3 \\ 
        Average\_WeekSel & 213 & 139 & 7 \\ 
        TBATS & 215 & 142 & 2 \\ 
        RNN & 215 & 142 & 2 \\ 
        FBML\_XGB\_S\_WeekSel & 215 & 144 & 0 \\ 
        StackFeat\_LR\_S\_WeekSel & 216 & 141 & 2 \\ 
        ETS & 219 & 139 & 1 \\ 
        Stack\_LR\_S\_Log\_General & 219 & 140 & 0 \\ 
        StackFeat\_LinR\_S\_Log\_WeekSel & 222 & 137 & 0 \\ 
        Average\_WeekFull & 223 & 136 & 0 \\ 
        Auto.ARIMA & 224 & 134 & 1 \\ 
        StackFeat\_LR\_PH\_Log\_WeekSel & 226 & 133 & 0 \\ 
        StackFeat\_LinR\_S\_WeekSel & 226 & 132 & 1 \\ 
        DHR-ARIMA & 230 & 128 & 1 \\ 
        NNET-AR & 230 & 128 & 1 \\ 
        Stack\_LinR\_PH\_Log\_WeekSel & 238 & 120 & 1 \\ 
        Stack\_LinR\_S\_Log\_General & 239 & 120 & 0 \\ 
        Naïve & 242 & 117 & 0 \\ 
        StackFeat\_LinR\_PH\_Log\_WeekSel & 242 & 116 & 1 \\ 
        Random Walk & 243 & 113 & 3 \\ 
        Stack\_LR\_PH\_Log\_WeekSel & 243 & 116 & 0 \\ 
        Theta & 245 & 113 & 1 \\ 
        StackFeat\_LR\_PH\_WeekSel & 250 & 107 & 2 \\ 
        StackFeat\_LinR\_PH\_WeekSel & 251 & 108 & 0 \\ 
        Snaïve & 253 & 103 & 3 \\ 
        StackFeat\_XGB\_S\_WeekSel & 262 & 97 & 0 \\ 
        Stack\_XGB\_S\_Log\_General & 265 & 93 & 1 \\ 
        STLM-AR & 268 & 91 & 0 \\ 
        StackFeat\_XGB\_S\_Log\_WeekSel & 273 & 86 & 0 \\ 
        Stack\_XGB\_S\_Log\_WeekFull & 274 & 84 & 1 \\ 
        Stack\_XGB\_S\_Log\_WeekSel & 286 & 73 & 0 \\ 
        Stack\_XGB\_PH\_Log\_WeekSel & 297 & 62 & 0 \\ 
        StackFeat\_XGB\_PH\_WeekSel & 308 & 51 & 0 \\ 
        StackFeat\_XGB\_PH\_Log\_WeekSel & 317 & 41 & 1 \\ \hline
    \end{tabular}
\caption{Comparison of Stack\_LR\_S\_Log\_WeekSel with M4 weekly winning method, and other benchmarks and variants across all series of M4 weekly dataset in terms of the number of times the proposed method wins, loses and ties with the respective comparison method on sMAPE.} 
\label{tab:win_loss}
\end{table}

\subsection{Computational Performance}
We have executed our experiments on different hardware due to their extensiveness and thus computational times are in general not comparable. To perform a comparison study of computational performance, we execute a subset of the experiments in a controlled environment, namely an Intel(R) Core(TM) i7 processor (2.6GHz) and 32GB of main memory.

\begin{sloppypar}
Table \ref{tab:computational_time} shows the computational times of the considered eleven baseline models and the model that we propose as a strong ensemble weekly forecasting model, Stack\_LR\_S\_Log\_WeekSel, which is the best method among all variants across all datasets. The reported computational times of Stack\_LR\_S\_Log\_WeekSel include both sub-model forecast calculation time and forecast combination time. 
\end{sloppypar}

From Table \ref{tab:computational_time}, we can see that unsurprisingly, comparatively less complex models such as na\"ive and sna\"ive show the lowest computation times. The three highly accurate base models in the WeekSel pool, namely TBATS, DHR-ARIMA, and RNN, show considerably higher computational costs compared with the fourth method, Theta, which executes quickly. Out of all base models, RNN usually shows a much larger execution time except for the M4 and Dominick datasets where DHR-ARIMA and Auto.ARIMA have the highest execution times, respectively. The proposed combination method, Stack\_LR\_S\_Log\_WeekSel requires to execute the 4 base models before computing the final forecasts, and therewith shows the highest computational time. However, the overhead over the sum of the base model forecasting times is small.

\begin{table}
		\centering\fontsize{9}{9}\rm
		\begin{tabular}{rrrrrrrr}
			\toprule
			\cmidrule{2-8}
			& \ M4 & NN5    & Kaggle  & Ausgrid   & Traffic & Solar & Dominick     \\\cmidrule{2-8}
			\addlinespace
			  Theta & 3 & 0.3 & 2 & 1 & 2 & 0.3  & 0.2\\ 
			  TBATS & 2904 & 235 & 1407 & 516 & 433 & 40  & 28\\ 
              DHR-ARIMA & 11828 & 165 & 1910 & 1987 & 3839 & 106 & 14 \\ 
              RNN & 9025 & 5400 & 7200 & 7036 & 8986 & 2700  & 4980\\ 
              ETS & 13 & 1 & 9 & 3 & 8 & 1  & 2\\ 
              Naïve & 1 & 0.2 & 2 & 1 & 2 & 0.2  & 0.05\\ 
              Snaïve & 1 & 0.2 & 2 & 1 & 2 & 0.2 & 0.06\\ 
              Random Walk & 1 & 0.3 & 3 & 1 & 2 & 0.4  & 0.09\\ 
              STLM-AR & 6 & 1 & 7 & 2 & 44 & 5  & 0.2\\  
              NNET-AR & 1233 & 9 & 74 & 34 & 36 & 3  & 29\\ 
              Auto.ARIMA & 625 & 1002 & 1124 & 1503 & 322 & 36 & 41690\\ 
              Stack\_LR\_S\_Log\_WeekSel & 23880 & 5921 & 10639 & 9660 & 13380 & 2876 & 5030\\ 
              \bottomrule
		\end{tabular}
		\caption{Computational times (in seconds) of the baseline models and Stack\_LR\_S\_Log\_WeekSel across all datasets. The computational times of Stack\_LR\_S\_Log\_WeekSel include both sub-model forecast calculation time and forecast combination time.}\label{tab:computational_time}
\end{table}

\subsection{Individual Model Contributions in Optimal Oracle Combination Approach}

We further conduct a study to explore the contributions of the individual base models to forecasting accuracy. We only consider our best model pool, WeekSel for this analysis.
For this, we assume an optimal oracle combination approach, where the base model with the lowest error (sMAPE or MASE) on the test set is identified for each series and selected for forecasting. This procedure is followed considering the four base models in WeekSel, to identify a lower bound of errors. Then, the same procedure is followed by subsequently removing one of the base models, and only using the three remaining ones. This way, we are able to identify base models that perform better than the others on certain series, and are not consistently dominated by other methods. Therewith, these models add to the diversity of forecasting methods.
Table \ref{tab:ablation_study} shows the results of the study. The results of the lower bound containing the four base models, which is by definition the best-performing method in this comparison, are highlighted in boldface, and the results of the worst performing combinations are italicized. The combination containing only local models, TBATS\_DHR-ARIMA\_Theta is the worst performing combination for M4, NN5, Kaggle web traffic, Ausgrid and Dominick datasets, and the second worst combination for the Traffic dataset across all error metrics. The combinations with a mixture of local and global models show an overall better performance compared with TBATS\_DHR-ARIMA\_Theta. Thus, we can conclude that the RNN adds diversity to the forecasting pool of methods. Using a mixture of local and global models provides better forecasts as a result of incorporating the strengths of both global and local models while mitigating the weaknesses of each other.

\begin{table}
		\centering\fontsize{9}{9}\rm
		\begin{tabular}{rrrrrrrr}
			\toprule
			\cmidrule{2-8}
			& \ M4 & NN5    & Kaggle  & Ausgrid   & Traffic & Solar & Dominick   \\\cmidrule{2-8}
			\addlinespace
			\multicolumn{8}{l}{\bf Mean sMAPE} \\
			\addlinespace
			  TBATS\_DHR-ARIMA\_Theta & \textit{5.99} & \textit{9.95} & \textit{25.84} & \textit{19.63} & 10.07 & 18.41 & \textit{14.49} \\ 
              TBATS\_Theta\_RNN & 5.69 & 9.40 & 24.61 & 19.59 & \textit{11.39} & 18.89  & 13.25\\
              Theta\_RNN\_DHR-ARIMA & 5.83 & 9.37 & 24.45 & 18.54 & 9.93 & \textit{20.17}  & 13.21\\
              TBATS\_DHR-ARIMA\_RNN & 5.71 & 9.37 & 24.79 & 18.39 & 9.93 & 18.43  & 13.18\\
              \textbf{TBATS\_DHR-ARIMA\_Theta\_RNN} & \textbf{5.47} & \textbf{9.18} & \textbf{24.06} & \textbf{17.84} & \textbf{9.83} & \textbf{18.41}  & \textbf{13.02}\\
			\bottomrule
			\addlinespace
			\multicolumn{7}{l}{\bf Median sMAPE} \\
			\addlinespace
		      TBATS\_DHR-ARIMA\_Theta & \textit{3.58} & \textit{9.66} & \textit{23.27} & \textit{14.85} & 7.05 & 17.78 & \textit{12.69} \\ 
              TBATS\_Theta\_RNN & 3.41 & 8.96 & 22.49 & 14.33 & \textit{8.74} & 18.02  & \textbf{11.39}\\
              Theta\_RNN\_DHR-ARIMA & 3.41 & 8.75 & 22.36 & 14.04 & 7.02 & \textit{19.60}  & 11.76\\
              TBATS\_DHR-ARIMA\_RNN & 3.42 & 8.84 & 22.68 & 13.95 & 7.05 & 17.78  & \textbf{11.39}\\
              \textbf{TBATS\_DHR-ARIMA\_Theta\_RNN} & \textbf{3.24} & \textbf{8.66} & \textbf{22.11} & \textbf{13.66} & \textbf{6.99} & \textbf{17.78}  & \textbf{11.39}\\
			\bottomrule
			\addlinespace
			\multicolumn{7}{l}{\bf Mean MASE} \\
			\addlinespace
			  TBATS\_DHR-ARIMA\_Theta & \textit{1.736} & \textit{0.745} & \textit{0.587} & \textit{0.805} & 0.853 & 0.884  & \textit{0.811}\\ 
              TBATS\_Theta\_RNN & 1.664 & 0.700 & 0.564 & 0.800 & \textit{1.020} & 0.908  & 0.731\\
              Theta\_RNN\_DHR-ARIMA & 1.648 & 0.699 & 0.562 & 0.758 & 0.843 & \textit{0.978}  & 0.737\\
              TBATS\_DHR-ARIMA\_RNN & 1.648 & 0.699 & 0.571 & 0.761 & 0.846 & 0.886  & 0.733\\
              \textbf{TBATS\_DHR-ARIMA\_Theta\_RNN} & \textbf{1.561} & \textbf{0.684} & \textbf{0.555} & \textbf{0.732} & \textbf{0.837} & \textbf{0.884} & \textbf{0.724} \\
			\bottomrule
			\addlinespace
			\multicolumn{7}{l}{\bf Median MASE} \\
			\addlinespace
		      TBATS\_DHR-ARIMA\_Theta & \textit{1.170} & \textit{0.708} & \textit{0.465} & \textit{0.646} & 0.676 & 0.874  & \textit{0.723}\\ 
              TBATS\_Theta\_RNN & 1.097  & 0.636 & 0.447 & 0.606 & \textit{0.855} & 0.892  & \textbf{0.621}\\
              Theta\_RNN\_DHR-ARIMA & 1.113 & 0.655 & 0.446 & 0.618 & 0.666 & \textit{0.957}  & 0.650\\
              TBATS\_DHR-ARIMA\_RNN & 1.059 & 0.655 & 0.448 & 0.601 & 0.669 & 0.874  & 0.674\\
              \textbf{TBATS\_DHR-ARIMA\_Theta\_RNN} & \textbf{1.043} & \textbf{0.620} & \textbf{0.437} & \textbf{0.571} & \textbf{0.664} & \textbf{0.874}  & \textbf{0.621}\\
			\bottomrule
		\end{tabular}
		\caption{Results of the optimal oracle combination study. The results of the lower bound method are highlighted in boldface. The results of the worst performing combinations are italicized.}\label{tab:ablation_study}
\end{table}

\begin{table}
		\centering\fontsize{9}{9}\rm
		\begin{tabular}{rrrrrrrr}
			\toprule
			\cmidrule{2-8}
			& \ M4 & NN5    & Kaggle  & Ausgrid   & Traffic & Solar & Dominick     \\\cmidrule{2-8}
			\addlinespace
			  TBATS & 0.712 & 0.408 & 0.006 & 0.219 & 0.226 & 0.042 & 0.0\\ 
              DHR-ARIMA & 0.009 & 0.042 & 0.0 & 0.0 & 0.110 & 0.003 & 0.0\\ 
              Theta & 0.044 & 0.049 & 0.0 & 0.0 & 0.0 & 0.950 & 0.321\\ 
              RNN & 0.238 & 0.477 & 0.976 & 0.642 & 0.633 & 0.0 & 0.676 \\ 
              \bottomrule
		\end{tabular}
		\caption{Base Model Weights Chosen by Stack\_LR\_S\_Log\_WeekSel}\label{tab:lasso_weights}
\end{table}
    
\subsection{Analysis of Combination Weights}    
    
Table \ref{tab:lasso_weights} shows the base model weights chosen by our best stacking-based forecast combination approach, Stack\_LR\_S\_Log\_WeekSel across all datasets. TBATS is the best performing base model for M4, and RNN is the best performing base model for the NN5, Kaggle web traffic and Dominick datasets. Stack\_LR\_S\_Log\_WeekSel has also assigned the highest weights for the corresponding base models with those datasets and it shows this model has the capability of identifying the best base models that should be used for combining. Furthermore, the method also fully discards some of the base models, when computing the forecasts for Kaggle web traffic, Ausgrid, Traffic, Solar and Dominick datasets. Combining only the best base models while discarding the less important ones is a major benefit of using lasso regression for combining forecasts.

\subsection{Applicability of the Proposed Method to Series with Other Frequencies}

\begin{sloppypar}
Though developed for weekly time series, our methodology is applicable more broadly to other types of series. For an illustration, we analyse the performance of Stack\_LR\_S\_Log\_WeekSel across datasets of other frequencies such as daily data. Table \ref{tab:results_daily} shows the performance of that model and its variants that use the same base model pool and meta-learning algorithm across two daily datasets, namely the NN5 daily dataset and the Kaggle web traffic daily dataset, where both datasets are the original daily versions of the weekly datasets we considered before aggregation. The best-performing models are highlighted in boldface.
\end{sloppypar}

\begin{table}
       	\centering\fontsize{9}{9}\rm
		\begin{tabular}{rrrrr}
			\toprule
			\cmidrule{2-5}
			& \ mean sMAPE & median sMAPE & mean MASE & median MASE   \\\cmidrule{2-5}
			\addlinespace
			\multicolumn{5}{l}{\bf NN5 Daily} \\
			\addlinespace
			  ETS & 21.57 & 20.35 & 0.860 & 0.810 \\
			  Theta & 21.93 & 20.51 & 0.885 & 0.838 \\ 
              TBATS & 21.11 & \textbf{19.56} & 0.858 & 0.834 \\ 
              DHR-ARIMA & 26.87 & 24.65 & 1.078 & 1.055 \\ 
              RNN & 22.03 & 20.65 & 0.867 & 0.825 \\ 
              StackFeat\_LR\_PH\_WeekSel & 22.39 & 21.86 & 0.905 & 0.879 \\ 
              StackFeat\_LR\_PH\_Log\_WeekSel & 23.83 & 22.36 & 0.962 & 0.943 \\ 
              Stack\_LR\_PH\_Log\_WeekSel & 22.78 & 22.14 & 0.912 & 0.909 \\ 
              StackFeat\_LR\_S\_WeekSel & \textbf{21.00} & 20.21 & \textbf{0.828} & \textbf{0.804} \\ 
              StackFeat\_LR\_S\_Log\_WeekSel & 21.57 & 20.84 & 0.860 & 0.845 \\ 
              Stack\_LR\_S\_Log\_WeekSel & 21.47 & 19.76 & 0.854 & 0.835 \\   
			\bottomrule
			\addlinespace
			\multicolumn{5}{l}{\bf Kaggle Daily} \\
			\addlinespace
		      ETS & 53.50 & 47.63 & 1.580 & 0.910 \\
			  Theta & 43.13 & 40.23 & 0.957 & 0.758 \\ 
              TBATS & 42.52 & 39.42 & 0.848 & 0.730 \\ 
              DHR-ARIMA & 44.03 & 40.86 & 0.952 & 0.758 \\ 
              RNN & 39.38 & 37.55 & 0.788 & 0.695 \\ 
              StackFeat\_LR\_PH\_WeekSel & 51.11 & 47.67 & 1.330 & 0.907 \\ 
              StackFeat\_LR\_PH\_Log\_WeekSel & 40.69 & 39.06 & 0.821 & 0.722 \\ 
              Stack\_LR\_PH\_Log\_WeekSel & 40.70 & 38.96 & 0.818 & 0.725 \\ 
              StackFeat\_LR\_S\_WeekSel & 44.58 & 41.61 & 0.952 & 0.776 \\ 
              StackFeat\_LR\_S\_Log\_WeekSel & \textbf{39.30} & 37.59 & 0.788 & \textbf{0.694} \\ 
              Stack\_LR\_S\_Log\_WeekSel & \textbf{39.30} & \textbf{37.50} & \textbf{0.787} & \textbf{0.694} \\ 
			\bottomrule
			\end{tabular}
		\caption{Results of our best model, Stack\_LR\_S\_Log\_WeekSel and its variants across NN5 and Kaggle web traffic daily datasets. The best performing models of each dataset on each error metric are highlighted in boldface.}\label{tab:results_daily}
\end{table}

Our proposed model, Stack\_LR\_S\_Log\_WeekSel shows the best performance on the Kaggle web traffic daily dataset compared with the performance of sub-models and other variants across all error metrics. On the NN5 daily dataset, another variant of the model,  StackFeat\_LR\_S\_WeekSel shows the best performance across all error metrics except median sMAPE. Hence, the general methodology is applicable and may serve as a good ensemble model more broadly and for data of other frequencies. 

\section{Conclusions and Future Research}
\label{sec:conclusion}

Many businesses and industries nowadays deal with weekly data and need accurate forecasts on those. However, the forecasting literature is currently lacking easy-to-use, accurate, and automated approaches dedicated to the forecasting of weekly series.
In this paper, we have conducted an analysis to identify a strong automated ensemble model suitable to forecast weekly time series, based on the most recent trends in forecasting. We have analysed two meta-learning architectures, FBML and stacking, with three meta-learning algorithms: linear regression, lasso regression, and XGBoost, as well as three different base model pools. Based on all considered model variants, the stacking approach which optimally combines the forecasts of four base models: RNN, Theta, TBATS and DHR-ARIMA using lasso regression provided overall the best forecasts across seven experimental datasets. 

Our best performing model has also significantly outperformed a series of benchmarks such as simple averaging, FFORMA and the current state-of-the-art weekly forecasting models: TBATS and DHR-ARIMA. In particular, the suggested lasso regression model ranks first among the original contenders of the M4 weekly competition, based on mean sMAPE. Furthermore, it is applicable to any weekly dataset irrespective of the series length as it does not require time series features during model training where, to calculate features with proper seasonal handling, the series length should be at least two seasonal cycles. Hence, we conclude that our weekly forecasting model can be used as an easy-to-implement and automatic, strong ensemble model in the weekly time series forecasting space. 

The success of this model encourages as future work to further analyse the approach with data of other frequencies, where our preliminary experiments have shown that this model can be a possible approach to obtain accurate forecasts for daily data. Furthermore, it is worth to explore the methodological changes that are required to scale this method to larger datasets. Some possible changes would be replacing TBATS and the RNN with other forecasting models such as LightGBM that scale better with larger datasets, and using XGBoost or LightGBM as the meta-learning algorithm instead of lasso regression.

\section*{Acknowledgements}
This research was supported by the Australian Research Council under grant DE190100045, a Facebook  Statistics  for  Improving  Insights  and  Decisions  research  award,  Monash  University  Graduate Research funding and the MASSIVE High performance computing facility, Australia.

\bibliographystyle{elsarticle-harv}

\bibliography{sample}

\end{document}